\documentclass[runningheads]{llncs}
\usepackage[T1]{fontenc}
\usepackage{graphicx}
\usepackage{amsmath}
\usepackage{amssymb}
\usepackage{hyperref}
\usepackage{adjustbox}

\usepackage{color}

\begin{document}
\title{Contrastive Concept Importance: Explaining Pairwise Class Decisions Through Automatically Extracted Concept Representations}
\titlerunning{Contrastive Concept Importance}
\author{Roel Visser\orcidID{0009-0006-3067-5545}\thanks{Corresponding author.} \and
Isaac Roberts\orcidID{0009-0005-3875-729X} \and
Barbara Hammer\orcidID{0000-0002-0935-5591}}
\authorrunning{R. Visser et al.}
\institute{Bielefeld University, Inspiration 1, 33619 Bielefeld, Germany
\email{rvisser@techfak.uni-bielefeld.de}
}
\maketitle
\begin{abstract}
Concept-based explanations are a prevalent way to explain the decisions of complex black-box methods through semantically meaningful, human interpretable concepts.
To attribute the contribution of such concepts to a model's decisions, feature attribution methods are used to quantify how strongly each concept contributes to a model output.
These attributions are typically computed for a single (output) class and therefore answer a non-contrastive ``why $P$?'' question.
In many situations, however, such as cases of misclassification, class confusion, and low-margin predictions, the more natural question to ask is ``why $P$ rather than $Q$?''.
We introduce \emph{contrastive concept importance}, which attributes the logit margin between a target class and a contrast, or foil, class to concepts in an automatically extracted visual concept basis.
The resulting scores are signed, indicating whether a concept supports the target over the foil or the foil over the target, and can be decomposed into target-logit and foil-logit effects.
This makes it possible to distinguish globally important concepts from concepts that specifically influence
a class-pair distinction, including whether their effect is shared, one-sided, or directly contrastive.
We evaluate our method both qualitatively and quantitatively on a range of ImageNet class pairs.
Our results show that \emph{contrastive concept importance} reveals class-pair specific model behavior that is not captured by standard concept importance alone, as well as capturing information on the semantic structure of the underlying ImageNet classes.

\end{abstract}
\section{Introduction}

Due to the inherent complexity and opacity of deep neural networks, their decisions can be difficult for humans to interpret \cite{poeta2023concept}. To address this issue a wide variety of explainability (XAI) methods has been proposed, among which concept-based explanations.
Concept-based explanations aim to describe neural network decisions in terms of semantically meaningful, human interpretable higher order attributes, or abstractions, rather than individual pixels or latent dimensions \cite{poeta2023concept}.
On such concepts \emph{feature attribution} methods are then applied to obtain explanations of the form ``why $P$?'', by measuring the contribution of each concept to a specific prediction \cite{Fel2023a}.
However, explanations are often more natural and informative to humans when they are contrastive in nature, asking ``why $P$ rather than $Q$?'', where $P$ is some fact to be explained and $Q$ is some alternative option, or foil \cite{Miller2019,Stepin2021}.
This is especially relevant in cases of misclassification, class confusion, low-margin predictions, or visually similar classes, where the aim is not only to identify concepts that are relevant to a class, but concepts that distinguish it from a plausible alternative.

In order to obtain such contrastive explanations for concept-based XAI methods, we propose \emph{contrastive concept importance} (CCI), a concept-level attribution method for pairwise class decisions.
Building on contrastive explanation frameworks, we attribute the logit margin between a target class and a foil class to concepts in an automatically extracted visual concept basis.
We evaluate our method qualitatively and quantitatively on a range of ImageNet-1k class pairs, comparing CCI and standard class wise concept importance.

\section{Background}

Post-hoc, automatically extracted concept methods \cite{poeta2023concept,Fel2023a} have recently gained traction because they reveal not only \textit{where} a model focuses, but also \textit{what} semantic features it relies on \cite{Fel2023,poeta2023concept}. These features, referred to as concepts, have been shown to be more human-interpretable than raw pixel attributions \cite{Fel2023}. Such methods are typically framed within the Dictionary Learning paradigm \cite{Fel2023a}, with the goal of sparsely representing a black-box model's embeddings as a linear combination of human-interpretable concepts. To attribute concept importance to a classifier's predictions, prior work has used methods such as TCAV \cite{kim2018interpretability}, Sobol indices \cite{Fel2023,lookatthevariance}, and gradient-based approaches \cite{Fel2023a}. While these methods can explain misclassifications \cite{Fel2023a}, they may overlook concepts tied to a classifier's uncertainty \cite{roberts2025conceptualizinguncertaintyconceptbasedapproach}. Our work advances upon this literature by attributing concept importance \textit{contrastively}, directly incorporating pairwise class decision boundaries.

Contrastive explanations are commonly understood as answers to questions of the form ``why $P$ rather than $Q$?'', where $P$ is the fact, e.g. a model prediction, to be explained and $Q$ is some alternative outcome, or foil
\cite{Miller2019,Stepin2021}.
This view is motivated by work in philosophy and the social sciences showing that human explanations are often contrastive in nature, rather than exhaustive accounts of all causal factors
\cite{Miller2019}.
Interestingly, already early successful concepts for feature relevance learning
such as matrix learning vector quantization \cite{DBLP:conf/ijcnn/BiehlHSSV15} are contrastive in nature, highlighting the dimensions which distinguish two classes rather than the most representative ones for a single class.

Several methods exist for producing contrastive explanations through some application of feature attribution, e.g. \cite{Akula2020,Dhurandhar2018,Jacovi2021}, or reasoning over concepts, e.g. \cite{Canizales2026}.
Jacovi et al.~\cite{Jacovi2021} define contrastive explanations for neural classifiers by projecting latent representations onto a class-contrastive subspace, which they apply to an NLP setting.
Their method preserves the latent components that separate a model prediction from an alternative label and then measures contrastive behavior under interventions on factors such as textual highlights of input features or manually annotated concepts.
Our work shares the same fact-vs-foil motivation, but rather than using a contrastive latent projection as the explanation, we use the target-vs-foil logit margin as a scalar quantity that is attributed over an automatically extracted unsupervised visual concept basis.
Conversely, Canizales et al.~\cite{Canizales2026} apply a method for abductive and contrastive reasoning over sets of instances given a dictionary of concepts, whereas our method performs local attribution over the concept basis for each individual instance.
Akula et al.~\cite{Akula2020} propose CoCoX, a method for obtaining counterfactual explanations through finding the minimal set of concepts, obtained through K-Means, which need to be removed or added in order for the prediction to change.

\section{Method}

We propose \emph{contrastive concept importance} as a method for concept-level explanations of pairwise model decisions. Rather than estimating the importance of a concept for a single class logit, we attribute the logit margin between a target and foil class to concepts extracted from an internal model representation. This makes the explanation explicitly pair-specific, meaning a concept is important when it changes the model's preference for one class over another.
Thus, providing a contrastive explanation "why P, rather than Q".

\subsection{Concept representation}
Let $\mathcal{D}=\{(x_i,y_i)\}_{i=1}^n$ denote a labelled dataset, where $(x_1,\ldots,x_n)\in \mathcal{X}^n \subseteq \mathbb{R}^{n\times d}$ are $n$ input images and $(y_1,\ldots,y_n)\in \mathcal{Y}^n$ are their associated labels.
We consider a trained classifier
$
f:\mathcal{X}\rightarrow \mathbb{R}^{C},
$
where $f(x)=\mathbf{z}(x)$ denotes the logit vector over $C$ classes.
Following CRAFT~\cite{Fel2023}, we decompose the trained classifier as $f = h \circ g$, where $g$ maps an input image to an intermediate activation vector $\mathbf{a}(x)=g(x)$ and $h$ maps this representation to the logit vector $\mathbf{z}=h(\mathbf{a}(x))$.
For a set of images or crops $\mathbf{X}$, we denote the corresponding activation matrix by $\mathbf{A} = g(\mathbf{X}) \in \mathbb{R}^{n \times p}$.
A concept basis is obtained by applying non-negative matrix factorization (NMF) to these activations, solving
$$
(\mathbf{U},\mathbf{W}) = \underset{{\mathbf{U}\geq0,\mathbf{W}\geq0}}{\arg\min} \frac{1}{2} \|\mathbf{A} - \mathbf{U}\mathbf{W}^\top\|_F^2,
$$
where $\mathbf{U} \in \mathbb{R}^{n \times r}$ contains the concept coefficients and $\mathbf{W} \in \mathbb{R}^{p \times r}$ contains the $r$ concept activation vectors (CAVs), or concept bank.
Each activation is therefore approximated as a non-negative combination of concept directions.
Once $\mathbf{W}$ has been fitted, the concept coefficient vector $\mathbf{u}(x)$ of a new input can be obtained by solving the corresponding non-negative least-squares projection onto the fixed concept bank.

\subsection{Contrastive concept attribution}

Given a target class $a$ and foil class $b$, we define the original logit margin as
$$
M_{a,b}(x)=z_a(\mathbf{a}(x))-z_b(\mathbf{a}(x)),
$$
where $z_k(x)$ denotes the logit for output class $k$.
Let $\mathcal{B}$ denote a fixed concept basis with concept bank $\mathbf{W}_{\mathcal{B}}$.
In our experiments, $\mathcal{B}$ is the target-class concept basis, but the definition also applies to a foil-class or shared basis.
We evaluate the margin as a function of an arbitrary concept vector $\mathbf{u}$ through the reconstructed activation
$
\hat{\mathbf{a}}_{\mathcal{B}}(\mathbf{u})=\mathbf{u}\mathbf{W}_{\mathcal{B}}^\top,
$
yielding the concept space margin
$$
\hat{M}_{a,b}^{(\mathcal{B})}(\mathbf{u})=z_a(\hat{\mathbf{a}}_{\mathcal{B}}(\mathbf{u}))-z_b(\hat{\mathbf{a}}_{\mathcal{B}}(\mathbf{u})).
$$
This scalar function is our contrastive attribution target, namely rather than attributing a single class logit, we attribute the model's preference for $a$ over $b$.

For an input $x$ with concept coefficients $\mathbf{u}_{\mathcal{B}}(x)$, we compute Integrated Gradients (IG)~\cite{Sundararajan2017} with respect to the concept coefficients.
IG is method for attributing the importance of features by accumulating the gradients along a path between a baseline and the actual input.
While \cite{Fel2023} employ SOBOL for attributing concept importance, which is variance based and provides nonnegative scores, we apply IG as it provides a more intuitive notion of directionality to the contribution to the margin, allowing for both positive and negative progression along the gradient from the baseline.
This gives more meaningful insight into how a concept influences the output by supporting or suppressing one class, the other, or both.
Furthermore, \cite{Fel2023a} found that IG is one of the methods which is optimal in terms of fidelity when used, as we do, in the penultimate layer.
Given a baseline concept vector $\mathbf{u}'_{\mathcal{B}}$, the attribution of concept $i$ to the target-vs-foil margin is
$$
\operatorname{CCI}^{(\mathcal{B})}_{i,a,b}(x)
=
\left(u_{\mathcal{B},i}(x)-u'_{\mathcal{B},i}\right)
\int_{0}^{1}
\frac{\partial \hat{M}^{(\mathcal{B})}_{a,b}
\left(\mathbf{u}'_{\mathcal{B}}+\alpha\left(\mathbf{u}_{\mathcal{B}}(x)-\mathbf{u}'_{\mathcal{B}}\right)\right)}
{\partial u_{\mathcal{B},i}}
d\alpha .
$$
We refer to this quantity as \emph{contrastive concept importance}.
In this work, we use the zero vector $\mathbf{u}'_{\mathcal{B}}=\mathbf{0}$ as the baseline concept activation, corresponding to the absence of all concepts in the basis.
For readability, we omit $(\mathcal{B})$ when the basis is fixed.
The sign of $\operatorname{CCI}_{i,a,b}(x)$ indicates the direction of the concept's effect on the class contrast.
The magnitude $|\operatorname{CCI}_{i,a,b}(x)|$ indicates how strongly the concept contributes to the boundary between the two classes.

We can also aggregate local explanations over a cohort of samples to obtain a global class-boundary profile.
For a cohort $\mathcal{S}$, such as samples predicted as the target class
or a subset of confused samples, we define the mean signed contrastive importance and the mean absolute contrastive importance as
$$
\Gamma_{i,a,b}(\mathcal{S})
=
\frac{1}{|\mathcal{S}|}
\sum_{x\in\mathcal{S}}
\operatorname{CCI}_{i,a,b}(x),
\qquad
\Gamma^{\mathrm{abs}}_{i,a,b}(\mathcal{S})
=
\frac{1}{|\mathcal{S}|}
\sum_{x\in\mathcal{S}}
\left|
\operatorname{CCI}_{i,a,b}(x)
\right|.
$$

\section{Experimental evaluation}

We evaluate CCI through target-vs-foil class pair analyses on ImageNet-1k.
The experiments are organized around two central claims.
First, contrastive attribution provides information beyond standard class-specific concept importance by identifying signed evidence for a class-pair distinction.
Second, the resulting scores correspond to meaningful model behavior, as tested through concept deletion, logit-wise decomposition, and semantic hierarchy comparisons.
Together, these analyses assess whether CCI is not merely a different ranking of concepts, but a useful tool for explaining how a model distinguishes between a target class and relevant foil.

For our experiments we use the ImageNet-1k dataset \cite{Deng2009ImageNet}.
Results are given for a ResNet50 model from the TIMM collection \cite{rw2019timm} (``nf\_resnet50.ra2\_in1k'') which
balances overall performance
with persistent confusion on more difficult or semantically similar class pairs.
Following \cite{Fel2023a,Fel2023}, we extract concepts from the penultimate activation layer.
We use class-specific concept bases with $r=10$ concepts
, which provides a compact concept basis while retaining enough semantically meaningful concepts needed to properly describe the model's predictions and behavior.
The methods and experiments are available at {\footnotesize{\url{https://github.com/r-visser/contrastive-concept-importance}}}.

\subsection{Example: beagle vs. English foxhound}
\begin{figure}[t]
\centering
\begin{tabular}{cc}
\includegraphics[width=0.335\textwidth]{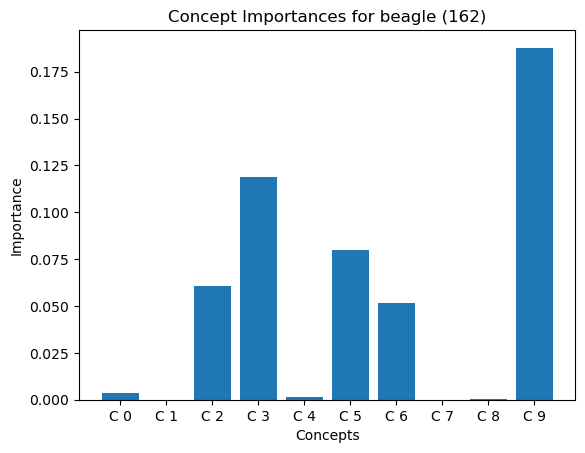}
&
\includegraphics[width=0.53\textwidth]{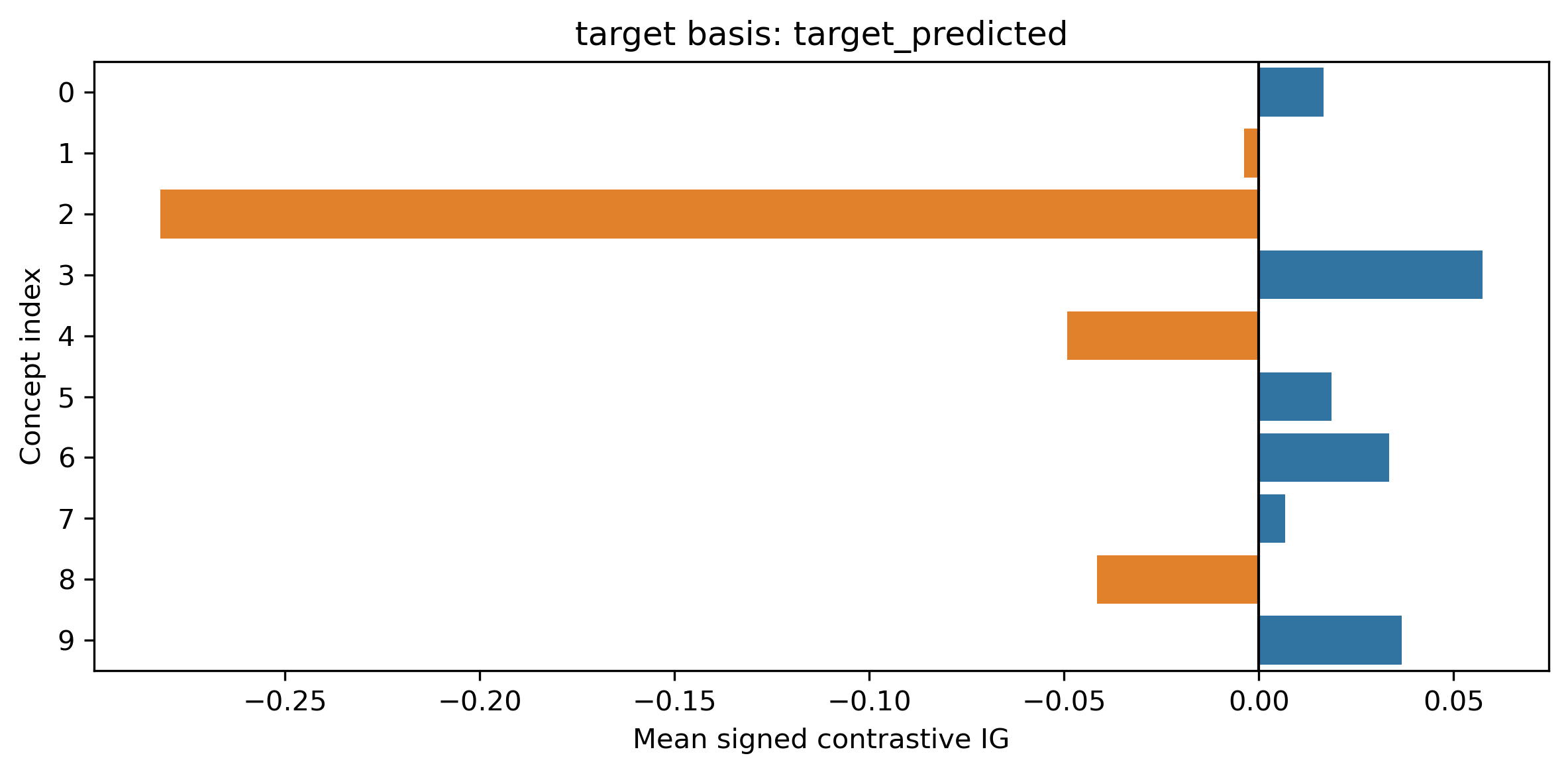}
\\
\includegraphics[width=0.42\textwidth]{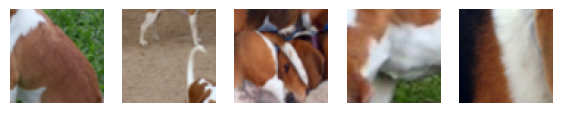}
&
\includegraphics[width=0.42\textwidth]{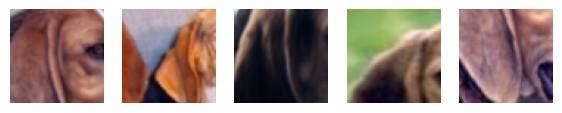}
\end{tabular}
\caption{
Example of regular and contrastive concept importance for the \emph{beagle} basis on the \emph{beagle-vs-foxhound} class pair.
The top-left panel shows class-wise global concept importance, as implemented by \cite{Fel2023}, for the \emph{beagle} class, while the top-right panel shows signed global CCI for the margin $M_{\mathrm{beagle},\mathrm{foxhound}}$ over the target-predicted cohort.
The bottom-left and bottom-right panels show example crops for concept~2 and concept~9, respectively.
}
\label{fig:beagle-foxhound-example}
\end{figure}

To showcase the method we first consider the fine-grained distinction between \emph{beagle} and \emph{English foxhound}.
This class pair is visually similar, belongs to the same hound/dog group of the ImageNet hierarchy, and is among the higher confusion pairs for the model used in our experiments.
Figure~\ref{fig:beagle-foxhound-example} compares global concept importance with global CCI for the \emph{beagle} concept basis.
Standard concept importance identifies concepts that are useful for the beagle prediction in general, but does not indicate whether those concepts help distinguish beagles from English foxhounds.
CCI instead ranks concepts by their signed effect on the \emph{beagle-vs-foxhound} margin, and can therefore separate concepts that support the target class from concepts that support the foil class for this specific decision.

Concept~2, for example, receives a strongly negative contrastive attribution.
This means that the concept decreases $M_{\mathrm{beagle},\mathrm{foxhound}}$ and therefore supports the foil class relative to the target class for this decision boundary.
This interpretation differs from the standard class importance view, where every important \emph{beagle} basis concept might be read as evidence for \emph{beagle} (globally or locally) but does not reveal how the model is actually distinguishing between classes to come to its decision.
The example illustrates the central distinction that standard concept importance explains class-specific evidence, whereas CCI explains evidence for the pairwise decision.
As well as showing how class-specific contributions can differ from those relevant to a class pair, seeing how e.g. concept~9 is much less important for the class pair distinction versus concept~2 which is much more relevant pairwise than it is globally.

\subsection{Concept interventions}
Concept interventions are a common tool for evaluating concept based explanation, used by e.g. \cite{Canizales2026,Fel2023a,Fel2023,Jacovi2021}.
For the following experiments interventions are used as an explanatory and validation tool, namely if a concept is important for a target-vs-foil decision, perturbing the corresponding concept (direction) should produce a corresponding, measurable change in the margin or downstream prediction behavior, such as pairwise confusion.
We have two paths for intervention.
Following the concept-space fidelity analyses from \cite{Fel2023}, we can modify the concept coefficients directly and then reconstruct the activations through NMF.
However, given the lossy nature of the reconstruction we also introduce a direct activation-space intervention where we perturb the original activation $\mathbf{a}(x)$ directly along the corresponding CAV direction $\mathbf{w}_{c,i}$.
This gives an activation-level probe of the same concept direction without requiring full reconstruction from the concept basis.

Let
$
\hat{\mathbf{w}}_{c,i}=\frac{\mathbf{w}_{c,i}}{\|\mathbf{w}_{c,i}\|_2}
$
denote the normalized CAV direction.
A removal intervention subtracts the projection of the activation onto this direction,
$$
\mathbf{a}^{\mathrm{as}}_{c,i}(x)
=
\mathbf{a}(x)
-
\left\langle \mathbf{a}(x), \hat{\mathbf{w}}_{c,i}\right\rangle
\hat{\mathbf{w}}_{c,i}.
$$
Analogous increase or decrease interventions can be defined by moving $\mathbf{a}(x)$ along $\hat{\mathbf{w}}_{c,i}$.
For removal of multiple CAVs simultaneously we can subtract them by projection onto the span of selected CAV directions.

\subsection{Fidelity of contrastive rankings}

\begin{figure}[t]
\centering
\begin{tabular}{cc}
\includegraphics[width=0.63\textwidth]{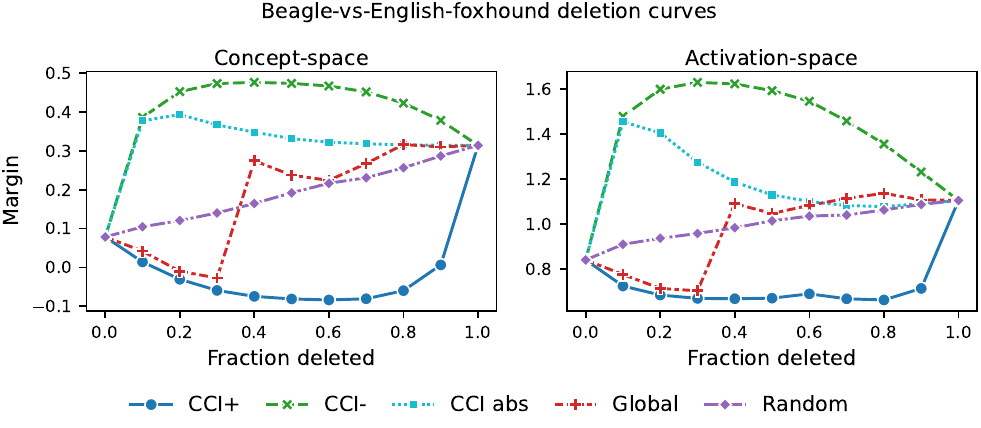}
&
\includegraphics[width=0.34\textwidth]{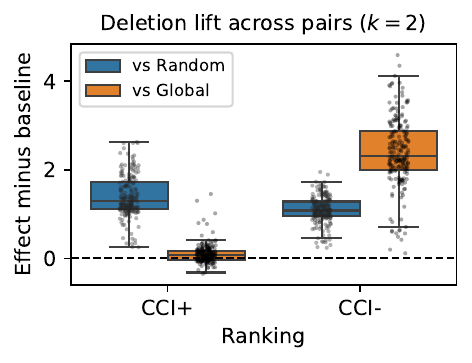}
\end{tabular}
\caption{
Deletion-curve fidelity for contrastive concept rankings.
Left: deletion curves for the \emph{beagle-vs-English-foxhound} class pair using the \emph{beagle} concept basis. The concept-space curve applies deletion through NMF reconstruction, while the activation-space curve removes concept by projection onto the span of selected CAV directions.
Right: aggregate activation-space deletion lift across 202 class pairs at $k=2$ concepts. Lift is the expected direction margin effect of the CCI ranking minus the corresponding effect of a random or class-wise global importance ranking. Positive values indicate that CCI ranked concepts affect the target-vs-foil margin more strongly than the baseline.
}
\label{fig:beagle-foxhound-deletion-fidelity}
\end{figure}

We evaluate whether the concept ranking induced by $\operatorname{CCI}$ is faithful to the model's target-vs-foil behavior using deletion curves.
Following \cite{Fel2023a,Fel2023}, we obtain activations from progressively smaller subsets of concepts and measure the resulting pairwise margin $M_{a,b}$.
A faithful contrastive ranking should affect this margin more systematically than global importance or a random ordering, and signed rankings should move the margin in the direction predicted by their attribution.

The deletion curves in Figure~\ref{fig:beagle-foxhound-deletion-fidelity} (left) show this ranking behavior for the \emph{beagle-vs-foxhound} class pair.
Concepts ranked by positive contrastive importance preserve or increase the \emph{beagle-over-foxhound} margin, whereas concepts ranked by negative contrastive importance move the margin toward the foil class.
This directional pattern is indeed not captured by standard global importance, which ranks concepts by their relevance to the beagle class but not by their effect on the \emph{beagle-vs-foxhound} distinction.
The contrastive ranking includes concepts that both support the target, the foil, or both.
The observed non-monotonicity is, therefore, expected for a signed target-vs-foil margin, highlighting precisely the directional behavior that CCI is intended to expose.
Because the CCI+ and CCI- rankings contains both positive and negatively signed concepts, turning points indicate that the deletion sequence has moved from removing target-supporting concepts to removing foil-supporting concepts, or vice versa.
The deletion curves therefore support that CCI identifies concepts that are faithful to the model's pairwise decision behavior, rather than merely producing a different ordering of class-important concepts.

Across 202 class pairs we summarize this effect (Figure~\ref{fig:beagle-foxhound-deletion-fidelity}, right) by measuring deletion lift, defined as the expected directional margin change of CCI ranking minus the corresponding change for a random or global importance ranking. The positive lift values show that CCI identifies concepts that are more important to the target-vs-foil distinction than random concepts and for CCI- concepts especially also than standard class-wise global concept importance.
Due to the lossy nature of the NMF projection,
it is important to note that we may need to be hesitant to use the full magnitude and absolute values of CCI score directly, but the fidelity experiment shows that concept rank and directionality information is preserved for both the concept-space projection path and original activation space, which can yield useful low dimensional explanations about between class prediction importance of concepts, especially when it's contrasted to standard concept importance.

\subsection{Directional logit-wise decomposition}
\label{sec:logit-decomposition}
While the view of contrastive importance in Figure \ref{fig:beagle-foxhound-example} give us a general sense of the influence of each concept on the pairwise class distinction, we can gain additional insight on their actual influence by looking at the specific role they play for each individual class.
By decomposing the contrastive attribution by attributing it to each logit individually,
we can visualize each concept's target-logit supporting component $P_i$, foil-logit suppression component $Q_i$, and overall contrastive concept importance $CCI_i=P_i+Q_i$.
This representation shows whether high-ranked concepts act through shared or non-contrastive effects on both logits, one-sided effects on one logit, or more directly contrastive effects that support one class while suppressing the other.

\begin{figure}[t]
\centering
\begin{tabular}{cc}
\includegraphics[width=0.5\textwidth]{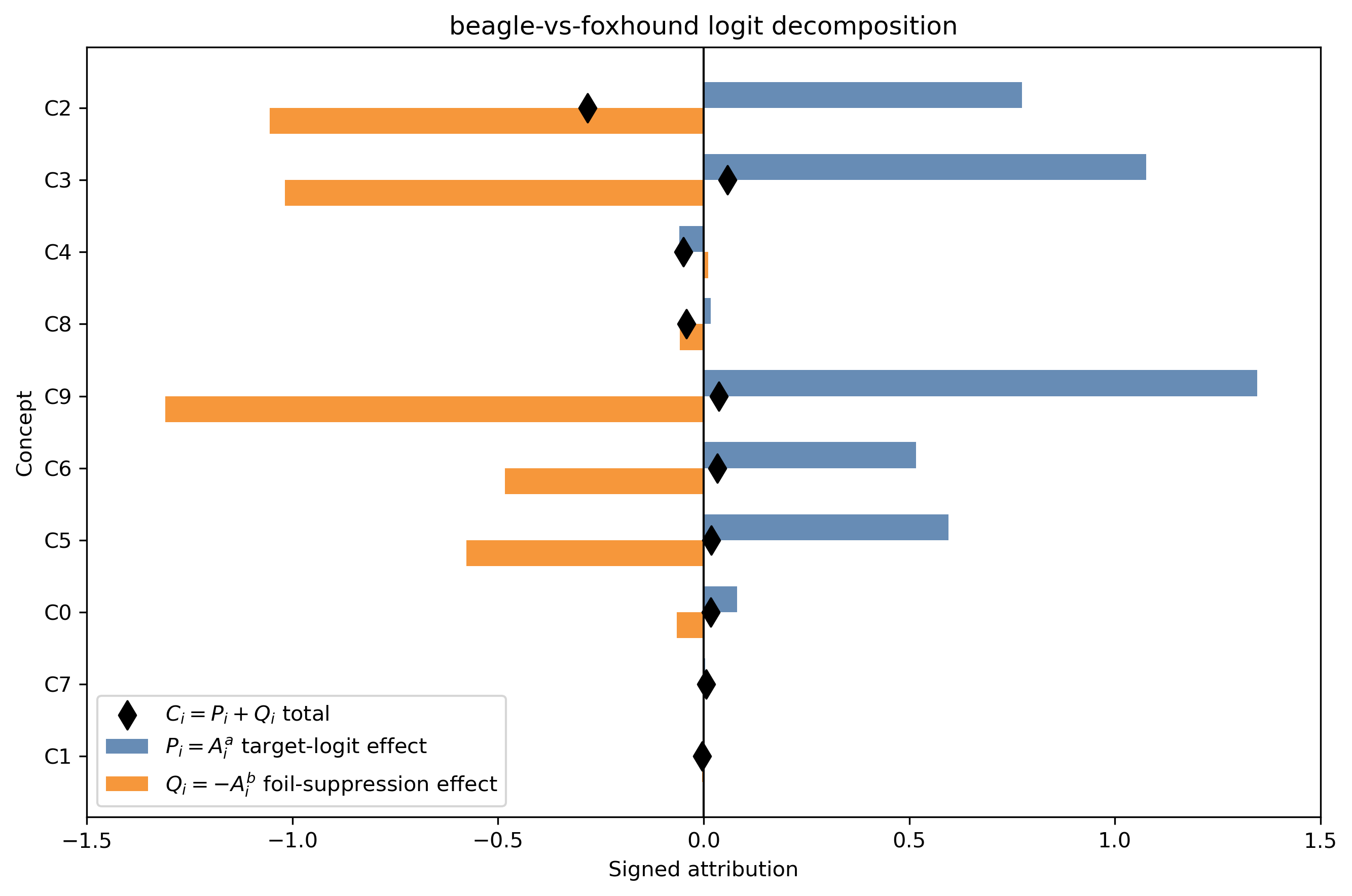}
&
\includegraphics[width=0.5\textwidth]{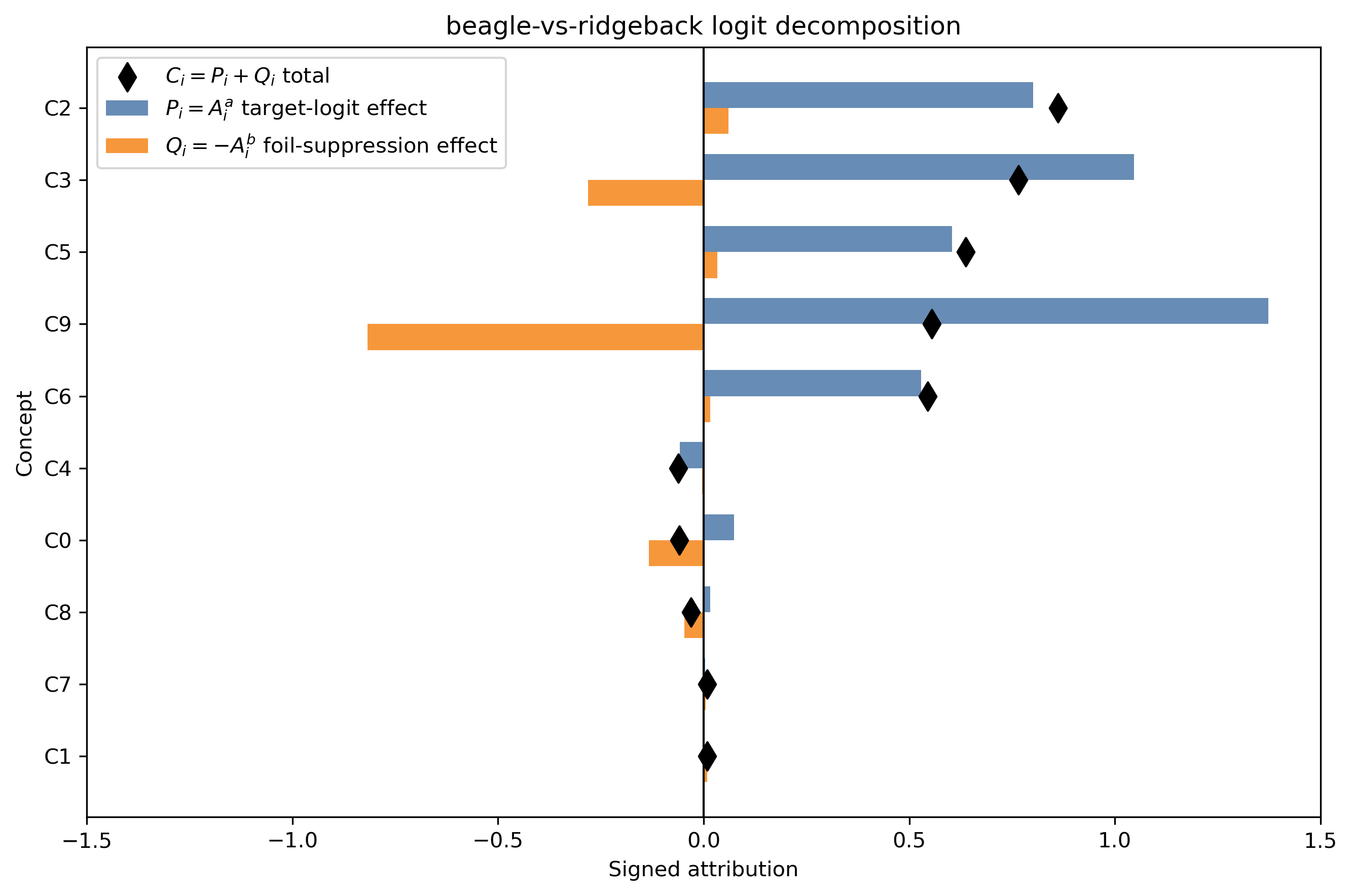}
\end{tabular}
\caption{
Logit-wise decomposition of contrastive concept importance for two beagle target-class contrasts.
The left panel contrasts \emph{beagle} with \emph{English foxhound}, a higher confusion fine-grained pair, while the right panel contrasts \emph{beagle} with \emph{Rhodesian ridgeback}, a lower confusion pair.
Each concept's contrastive contribution $CCI_i$ is decomposed into target-logit support $P_i$ and foil-logit suppression $Q_i$, showing whether the concept acts as shared evidence, one-sided evidence, or a more directly contrastive component.
}
\label{fig:logit-decomposition-pair-comparison}
\end{figure}

Figure~\ref{fig:logit-decomposition-pair-comparison} compares this decomposition for two \emph{beagle} target class contrasts.
For the high confusion \emph{Beagle} vs \emph{English Foxhound} pair, several important concepts have large target-logit effects but also large foil logit effects in the opposite direction of the contrast.
This indicates shared (hound- or dog-like) evidence that activates both classes and partly cancels in the margin.
For the lower confusion \emph{Beagle} vs \emph{Rhodesian ridgeback} pair, the same \emph{beagle} concept basis shows clearer target-supporting behavior.
Concept~9 remains an example of a globally important concept with a shared component, but its target logit contribution dominates more strongly for the \emph{ridgeback} contrast than for the \emph{foxhound} contrast.
Moreover, several concepts increase the \emph{beagle} logit while only weakly activating or even mildly suppressing the foil logit, leading to larger positive contrastive contributions across a wider set of concepts. Meaning that there is a larger amount of concepts which can be used as evidence for contrasting the target from the foil class.
While, conversely, the higher confusion pair has less means for contrasting the options, which may help explain the model's propensity for confusing these two classes.
The comparison suggests that CCI does not only rank
concepts by contrastive contribution
, but also reveals how the model represents easier and harder class distinctions differently. Providing potentially useful insight into why certain class pairs exhibit such high confusion or high uncertainty.

\subsection{Semantic hierarchy specificity}
As a further quantitative evaluation of our method we employ the semantic class hierarchy of the ImageNet-1k dataset \cite{Deng2009ImageNet}.
The goal of this experiment is to move beyond a single pairwise example and test whether concepts selected by contrastive importance are specific to the fine-grained target-vs-foil distinction, rather than merely reflecting broader superclass evidence.
For a fixed target class and target concept basis, we construct sweeps in which the foil class is moved progressively farther away in the semantic hierarchy.
For broad coverage we evaluate sweeps for 12 target classes from a wide range of categories across ImageNet's semantic tree.

ImageNet-1k was constructed according to the WordNet lexical database \cite{Deng2009ImageNet}.
Every class corresponds to a keyword description of a meaningful concept.
The concepts in WordNet are structured according to a semantic tree structure, wherein the ImageNet classes are leaf nodes that are categorized according to semantic categories such as vehicles, mammals, birds, dogs, etc.
This structure allows us to compare the between-class contrast, such as \emph{beagle} versus \emph{English foxhound}, to broader group contrasts, such as dog-vs-rest, mammal-vs-rest, or organism-vs-rest.

For each target-vs-foil pair we intervene on the top-2 ranked contrastive concepts and measure two quantities.
The fine effect is the absolute change in the pairwise class margin,
$
\Delta_{\mathrm{fine}}
=
\left|
\Delta M_{a,b}(x)
\right|
$
The hierarchy effect is the absolute change in a broader semantic group contrast,
$
\Delta_{\mathrm{group}}
=
\left|
\Delta M_{G,\mathrm{rest}}(x)
\right|,
$
where $G$ is a reference synset, such as dog or vehicle, and the rest contrast compares this group against ImageNet classes outside the group.
For a set of class indices $S$, we define the group score by log-sum-exp aggregation,
$
Z_S(x)
=
\log \sum_{k\in S}\exp(z_k(x)).
$
The rest group margin is then
$
M_{G,\mathrm{rest}}(x)
=
Z_G(x)-Z_{G_{\mathrm{rest}}}(x),
$
where $G_{\mathrm{rest}}$ denotes the complement of the group $G$ in the ImageNet label set.
We then compute a bounded specificity score,
$$
S_{\mathrm{spec}}
=
\frac{
\Delta_{\mathrm{fine}}
}{
\Delta_{\mathrm{fine}}+\Delta_{\mathrm{group}}+\epsilon
}.
$$

\begin{figure}[t]
\centering
\begin{tabular}{cc}
\adjustbox{valign=c}{
\includegraphics[width=0.58\textwidth]{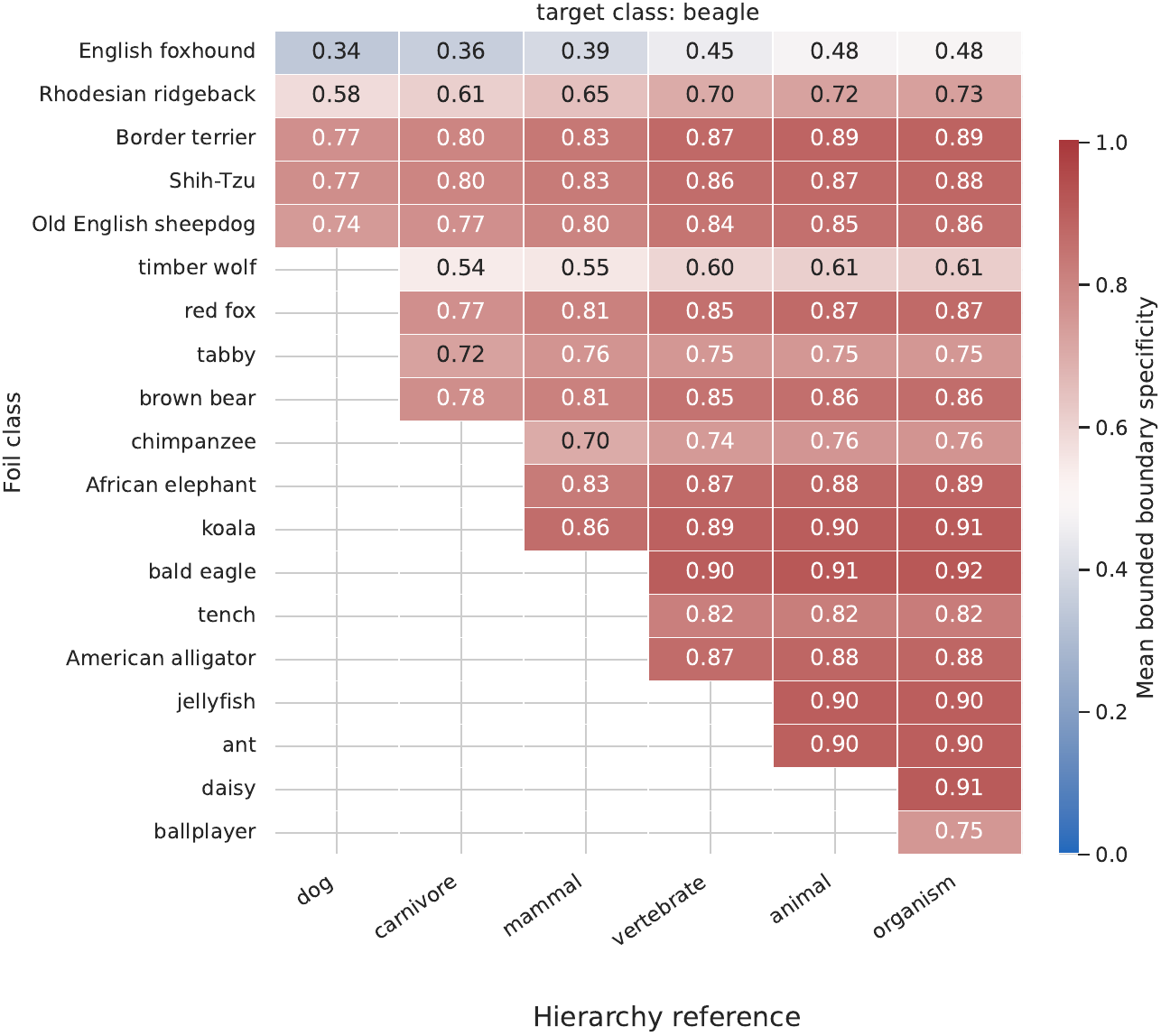}
}
&
\adjustbox{valign=c}{
\includegraphics[width=0.38\textwidth]{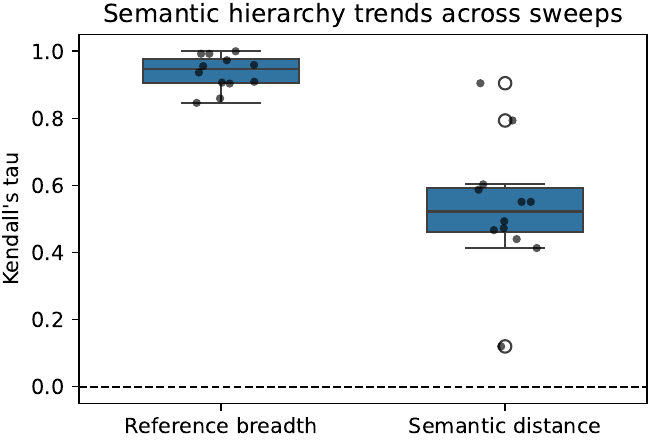}
}
\end{tabular}
\caption{
Semantic hierarchy specificity evaluation.
Left: specificity sweep for the \emph{beagle} target class using rest contrasts for the hierarchy references. Rows correspond to target-vs-foil class pairs at increasing semantic distance, and columns correspond to increasingly broad hierarchy group references.
High values indicate that interventions on contrastive concepts affect the fine target-vs-foil distinction more than the corresponding broader semantic group contrast.
Values near $0.5$ indicate comparable effects on the fine and group contrasts, while low values indicate that the intervention primarily changes broader superclass evidence.
Right: Kendall's $\tau$ rank correlations summarizing the same monotonic hierarchy pattern across multiple target-class sweeps. Positive values indicate that specificity tends to increase for broader hierarchy references and for more semantically distant foil classes.
}
\label{fig:semantic-hierarchy-rest-sweeps}
\end{figure}

Figure~\ref{fig:semantic-hierarchy-rest-sweeps} evaluates whether boundary specificity follows the semantic hierarchy.
For semantically close pairs, such as \emph{beagle-vs-foxhound}, the specificity is lower at more semantically close group levels (e.g. \emph{dog} and \emph{carnivore}) indicating that interventions on highly contrastive concepts also affect shared (e.g. ``dog-like'') evidence.
This is in line with our analysis of the logit-wise decomposition comparison in Section~\ref{sec:logit-decomposition}, where we observed that the \emph{beagle-vs-foxhound} pair shows a high level of shared, or overlapping, contribution within the concepts. Highlighting how these concepts represent much of the ``dogness'' of the two classes, rather than properly distinguishing between the two.
By contrast, as the foil becomes semantically more distant, the intervention becomes increasingly specific to the fine target-vs-foil distinction.
This produces the expected heatmap pattern, where specificity increases for broader hierarchy reference groups and also tends to increase as the foil moves farther away from the target class, becoming more semantically distant.
Across sweeps, the Kendall's $\tau$ coefficients are consistently positive, indicating that this monotonic pattern is not limited to the displayed target class.
This indicates that our method can meaningfully capture semantically relevant model behavior across semantic groups and concept bases.

\section{Conclusion}

We introduced CCI as a concept-level attribution method for explaining pairwise class decisions.
By attributing the logit margin between a target and foil class, the method identifies concepts that are not only important for a class, but specifically influence the model's preference for one class over another.
The resulting signed scores, together with their logit-wise decomposition, provide a compact way to inspect whether concepts act as shared or non-contrastive, one-sided, or more directly contrastive evidence.
Our experiments on ImageNet class pairs show that this contrastive view provides information that is not captured by standard concept importance explanations alone.
While our fidelity and semantic hierarchy experiments show that our method captures meaningful semantic and behavioral information across a broad range of different class pairs.

Our current work uses class-specific concept bases, which provide fine-grained class-level concepts but do not define a single shared dictionary over which all classes can be compared.
Future work should therefore evaluate shared concept dictionaries, such as those produced from sparse autoencoders, and additionally extend CCI to uncertainty-oriented settings such as reject options or conformal prediction sets.

\bibliographystyle{splncs04}
\bibliography{bibliography}

\end{document}